\documentclass[final,5p,times, twocolumn]{elsarticle}

\usepackage{hyperref}
\usepackage{lineno}
\usepackage{comment}
\usepackage{amsmath}
\usepackage{float}
\usepackage{graphicx}
\usepackage{svg}
\usepackage{pdfpages} 
\modulolinenumbers[5]
\usepackage{pgfplots}
\usepackage{amssymb}
\usepackage{multirow}
\usepackage{ gensymb }
\usepackage{multicol}
\usepackage{makecell}
\usepackage{caption}
\usepackage{fixltx2e}

\makeatletter
\g@addto@macro{\UrlBreaks}{\UrlOrds}
\makeatother

\usepackage{subfig}
\usepackage{caption}
\journal{arXiv} % !!!
\begin{document}
\begin{frontmatter}

\title{ \huge Deep Graph Convolutional Networks for Wind Speed Prediction}

\author{Tomasz Stańczyk $^{a}$} 
\ead{t.stanczyk@student.maastrichtuniversity.nl}

\author{Siamak Mehrkanoon $^{a,}$\corref{cor1}}
\ead{siamak.mehrkanoon@maastrichtuniversity.nl}

\cortext[cor1]{Corresponding author}

\address{$^{a}$Department of Data Science and Knowledge Engineering, Maastricht University, The Netherlands}

\begin{abstract}

Wind speed prediction and forecasting is important for various business and management sectors. In this paper, we introduce new models for wind speed prediction based on graph convolutional networks (GCNs). Given hourly data of several weather variables acquired from multiple weather stations, wind speed values are predicted for multiple time steps ahead. In particular, the weather stations are treated as nodes of a graph whose associated adjacency matrix is learnable. In this way, the network learns the graph spatial structure and determines the strength of relations between the weather stations based on the historical weather data. We add a self-loop connection to the learnt adjacency matrix and normalize the adjacency matrix. We examine two scenarios with the self-loop connection setting (two separate models). In the first scenario, the self-loop connection is imposed as a constant additive. In the second scenario a learnable parameter is included to enable the network to decide about the self-loop connection strength. Furthermore, we incorporate data from multiple time steps with temporal convolution, which together with spatial graph convolution constitutes spatio-temporal graph convolution. We perform experiments on real datasets collected from weather stations located in cities in Denmark and the Netherlands. The numerical experiments show that our proposed models outperform previously developed baseline models on the referenced datasets. We provide additional insights by visualizing learnt adjacency matrices from each layer of our models.

\end{abstract} 

\end{frontmatter}
%\linenumbers

%%%%%%%%%%%%%%%%%%%%%%%%%%%%%%%%%%%%%%%%
%%%%%%%%%% General Outline %%%%%%%%%%%%%
%%%%%%%%%%%%%%%%%%%%%%%%%%%%%%%%%%%%%%%%
\section{Introduction}\label{section_introduction}
Accurate weather forecasting is important for agriculture, transportation and management as it influences decision making processes and optimal settings in these sectors. Weather prediction can also be used to predict natural disasters, e.g. heatwaves or hurricanes and therefore contribute to saving human lives. Moreover, accurate weather elements prediction can shed light in analysis regarding the climate change \cite{oneill@2017}.

Previous approaches involve using numerical weather prediction (NWP) \cite{lorenc@1986}. NWP uses physical assumptions about the weather elements and mathematically models the atmosphere as a fluid. Weather variables are then predicted by means of simulation of partial differential equations \cite{richardson@2007}. However, this approach requires high computing power and processing the data might take up to several hours \cite{bauer@2015}. Furthermore, due to the weather assumptions mentioned above as well as the noise present in the data, the accuracy of NWP can degrade over time \cite{tolstykh@2005}. Machine Learning data driven based models on the other hand, do not make any assumptions about the atmosphere. Instead, these data driven based models take historical weather data as an input and train a model aiming to predict the target values for some time step ahead as an output. 

In recent years literature has witnessed the success of deep machine learning models in many application domains 
\cite{lecun@2015, krizhevsky@2012, bengio@2013, webb@2018, mehrkanoon@2018, mehrkanoon_kernel_blocks@2019, mehrkanoon2019cross, mehrkanoon2017scalable, mehrkanoon2017regularized}. 
In particular, deep Learning based models learn their own features during the training, so that the features are the most optimal for network and they do not require to be hand-crafted \cite{lecun@2015}. Deep Learning has already been used for predicting climate data \cite{scher@2018, salman@2015, fernandez2020deep}. Further, deep Learning methodologies based on Convolutional Neural Networks (CNNs) have already been successfully applied to the weather forecasting problem \cite{mehrkanoon2019deep2,trebing2020wind}. However, CNN-based approaches do not incorporate spatial relations between the weather stations. The authors in \cite{mehrkanoon2019deep2}, casted the historical data in a tensorial format (weather stations, weather variables, time steps) which was then passed to the model and convolution operation was performed over the data volume. In this way, the neighborhood relation between the weather stations is only based on their order in the dataset.

Graph Convolution Networks (GCNs) can generalize CNNs to work on graphs rather than on regular grids \cite{zhang@2019}. In particular, it enables incorporating the neighbor relation information, e.g. through an adjacency matrix of a graph. GCNs have already been applied to various domains, e.g. computer vision \cite{johnson@2018, litany@2017}, natural language processing \cite{peng@2018, zhangdeng@2019} or natural sciences \cite{fout@2017, xie@2018}. The application of GCNs to weather prediction has not been yet extensively discovered.

In this work, we treat weather stations and their corresponding weather variable values from different time steps as a spatio-temporal graph, as presented in Fig \ref{fig:denmark_graph}(b). Here, we develop our own novel models from ST-GCN \cite{yan@2018} and 2s-AGCN \cite{shi@2018} architectures which were successful on skeleton-based action recognition tasks. Our models take as an input tensor data containing values of different weather variables (e.g. temperature, wind speed, air pressure among the others) for several cities and historical time steps and return values wind speed for selected cities and for selected time step ahead. Furthermore, the adjacency matrix containing the values of relations between the cities is optimized together with other parameters during the training phase. In this way, the network is able to learn and decide about the city relations based on the given training data.

The proposed models are trained on two datasets containing the weather data from Danish and Dutch datasets. These data have been previously introduced in \cite{mehrkanoon2019deep2} and \cite{trebing2020wind}. We evaluate our models using Mean Absolute Error and Mean Squared Error, which are the original metrics used for evaluation of the baseline models \cite{mehrkanoon2019deep2, trebing2020wind}. The obtained results show that the performances of our models surpass those of the baseline models. Prediction plots and error values are provided. Furthermore, since the learnable adjacency matrix is the only element deciding about the relations between the cities, we visualize what network has learnt and provide some interpretation insights.

This paper is organized as follows. Related work is provided in Section \ref{section_related_work}. Section \ref{section_methods} presents the proposed models and relevant details. Section \ref{section_experiments} describes the conducted experiments with corresponding results and Section \ref{section_discussion} includes discussion. Finally, conclusions are drawn in Section \ref{section_conclusion}.

\section{Related Work}\label{section_related_work}
Weather forecasting has already been approached by deep learning based methodologies. Although Long Short-Term Memory (LSTM) recurrent networks \cite{hochreiter@1997}, work well for time series prediction \cite{siaminamini@2018}, they do not include spatial relations within the data. Authors in \cite{shi@2015} combined LSTM with convolutions to create ConvLSTM, which was used for precipitation tasks, while also capturing the data spatial structure. The authors in \cite{mehrkanoon2019deep2} introduced models based on 1D-CNNs for multi-source 1D data, 2D- and 3D-CNNs to process the tensor-form 3D data. Spatial-temporal relations were extracted with 2D and 3D CNNs. The authors in \cite{trebing2020wind} proposed using depthwise-separable convolutions to process different dimensions of the input tensor and then concatenating resulting tensor along single dimension. However, all the CNN-based models above discard the spatial relations between the cities (weather stations). In other words, these approaches process input tensor, where neighborhood of the cities is determined only by the order of the cities in the tensor or the dataset.

GCN based approaches on the other hand can process non-grid data \cite{zhang@2019} and therefore include spatial relations between weather stations forming a graph. The authors in \cite{wilson@2018} introduced weighted graph convolutional LSTM architecture, which combines LSTM with matrix multiplications replaced with graph convolutions with a single (one for the whole model), learnable adjacency matrix. In \cite{wilson@2018} temperature and wind speed values were predicted. In \cite{seo@2017}, the authors proposed an architecture which combined latent representation of a graph coming from the encoder with output from an LSTM structure and temperature values were predicted. 
In \cite{khodayar@2019}, the authors created a graph based on the wind farms and for each node of the graph, temporal features were extracted with LSTMs. Their Graph Convolutional architecture was used for predicting wind speed.

In \cite{yu@2018} the authors introduced spatio-temporal graph convolutional networks (STGCN) to perform traffic forecasting. GCNs were used to extract spatial features and gated CNNs were used to extract temporal features. In \cite{yan@2018}, the authors introduced spatial temporal graph convolutional networks (ST-GCN, not to be confused with already mentioned STGCN) for skeleton-based action recognition. Adopted version of graph convolution from \cite{kipf@2017} was used as spatial convolution and standard 2D convolution as temporal convolution. The authors in \cite{yan@2018} used fixed graph with importance weighting. Further, authors in \cite{shi@2018} expanded the idea of ST-GCN and created a two-stream architecture with adaptive graph convolutional networks (2s-AGCN). Similar approaches as for ST-GCN were applied, yet with enabling the adjacency matrix to be partially fixed and partially learnable, and with no transformations included from \cite{kipf@2017}. The discussed approaches, using graph convolutional methodologies, reached remarkable performance in their target domains \cite{yu@2018, yan@2018, shi@2018}.

Based on the methodologies 
discussed above, we develop new novel models and apply them to the task of weather forecasting to predict wind speed in the selected cities (weather stations) in Denmark and The Netherlands. The proposed models are presented in Section  \ref{section_methods}.

\section{Methods} \label{section_methods}
\begin{figure*}[!htbp]
\centering
    \subfloat[]{\includegraphics[scale=0.325]{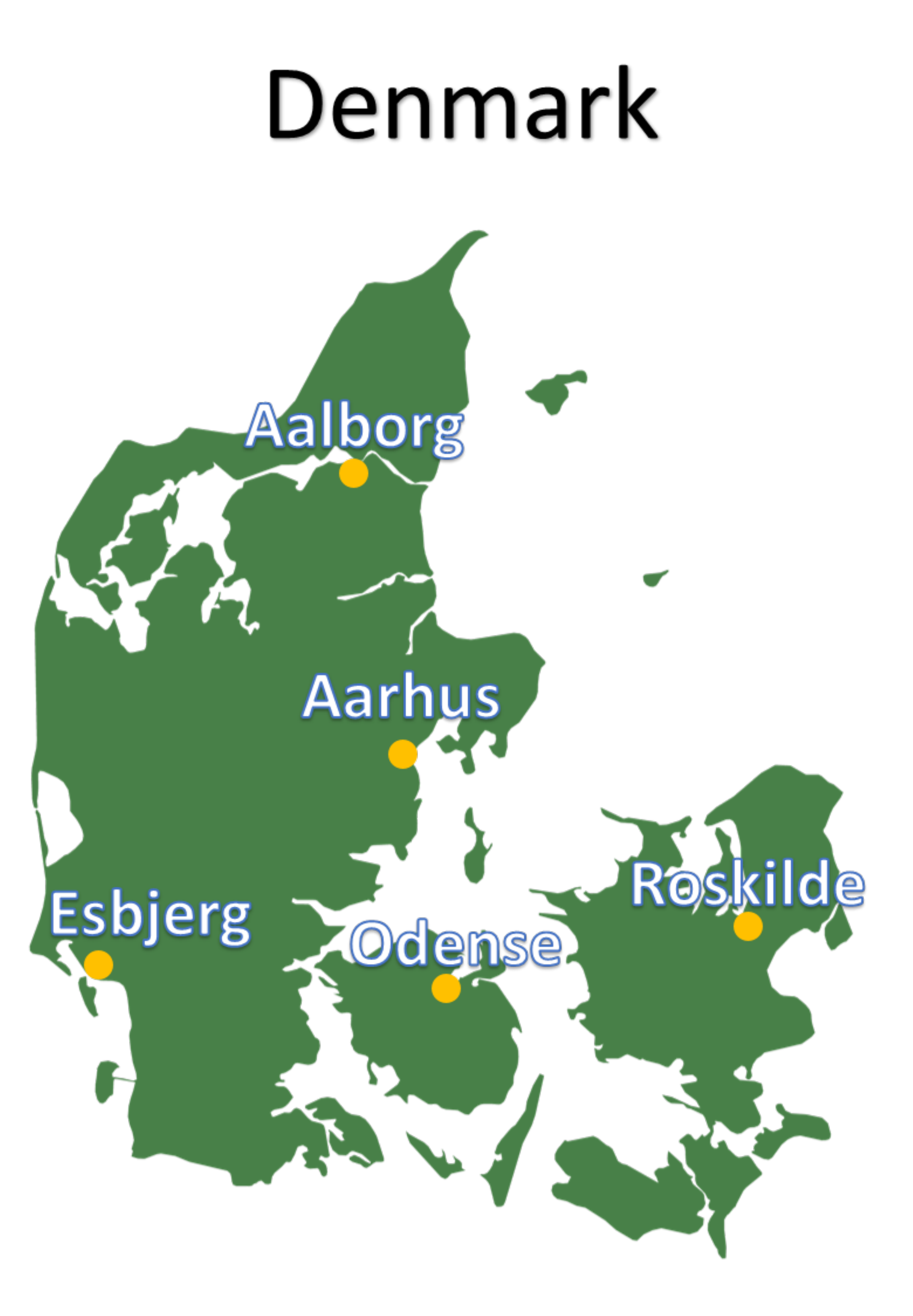}}
    \hspace{0.4cm}
    \subfloat[]{\includegraphics[scale=0.4]{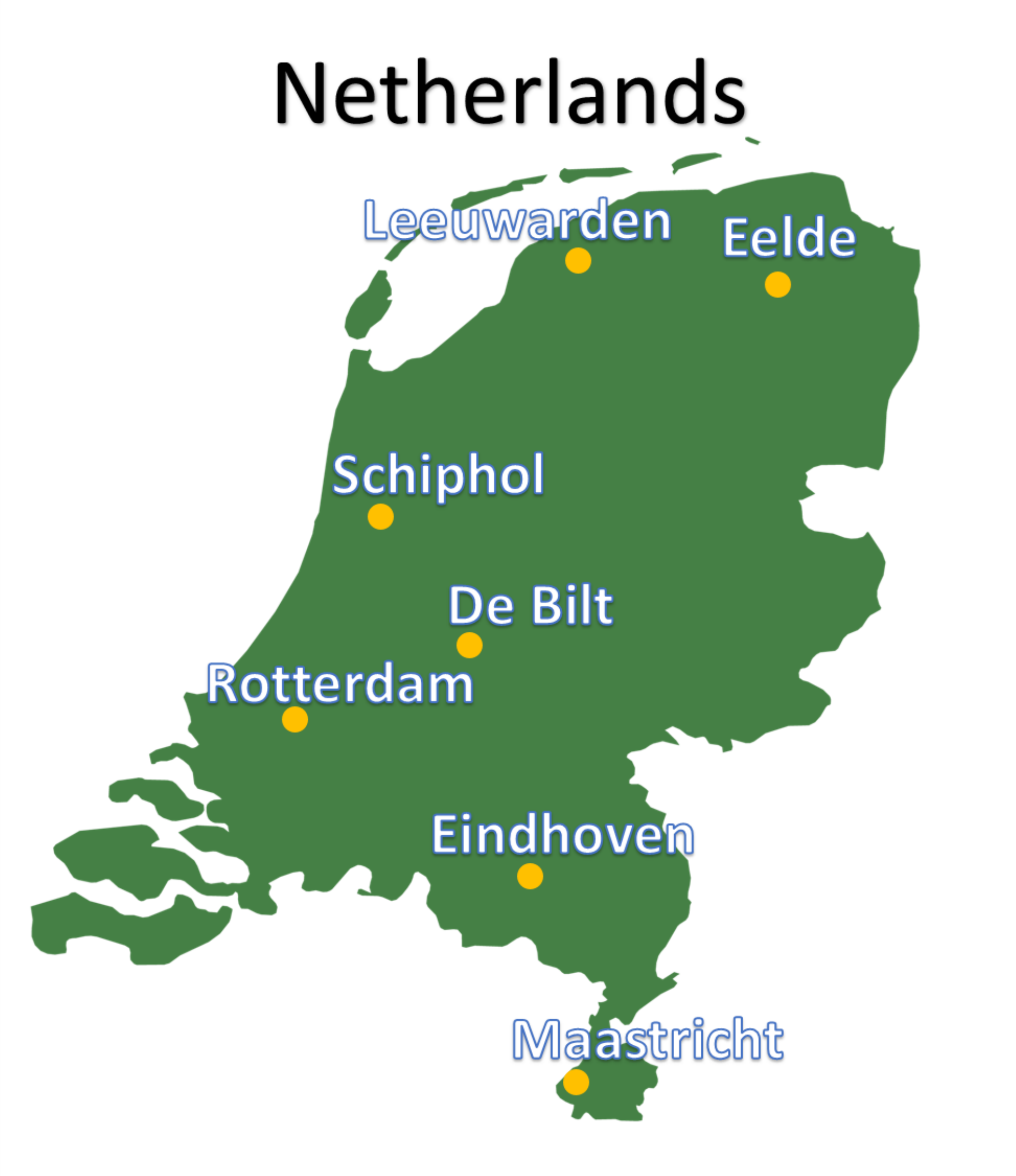}}
    \caption{Map with the weather stations including the data measured for the processed datasets: (a) Danish cities, (b) Dutch cities.}
    \label{fig:map_dk_and_nl}
\end{figure*}

\begin{figure*}[!htbp]
\centering
    \subfloat[]{\includegraphics[scale=0.32]{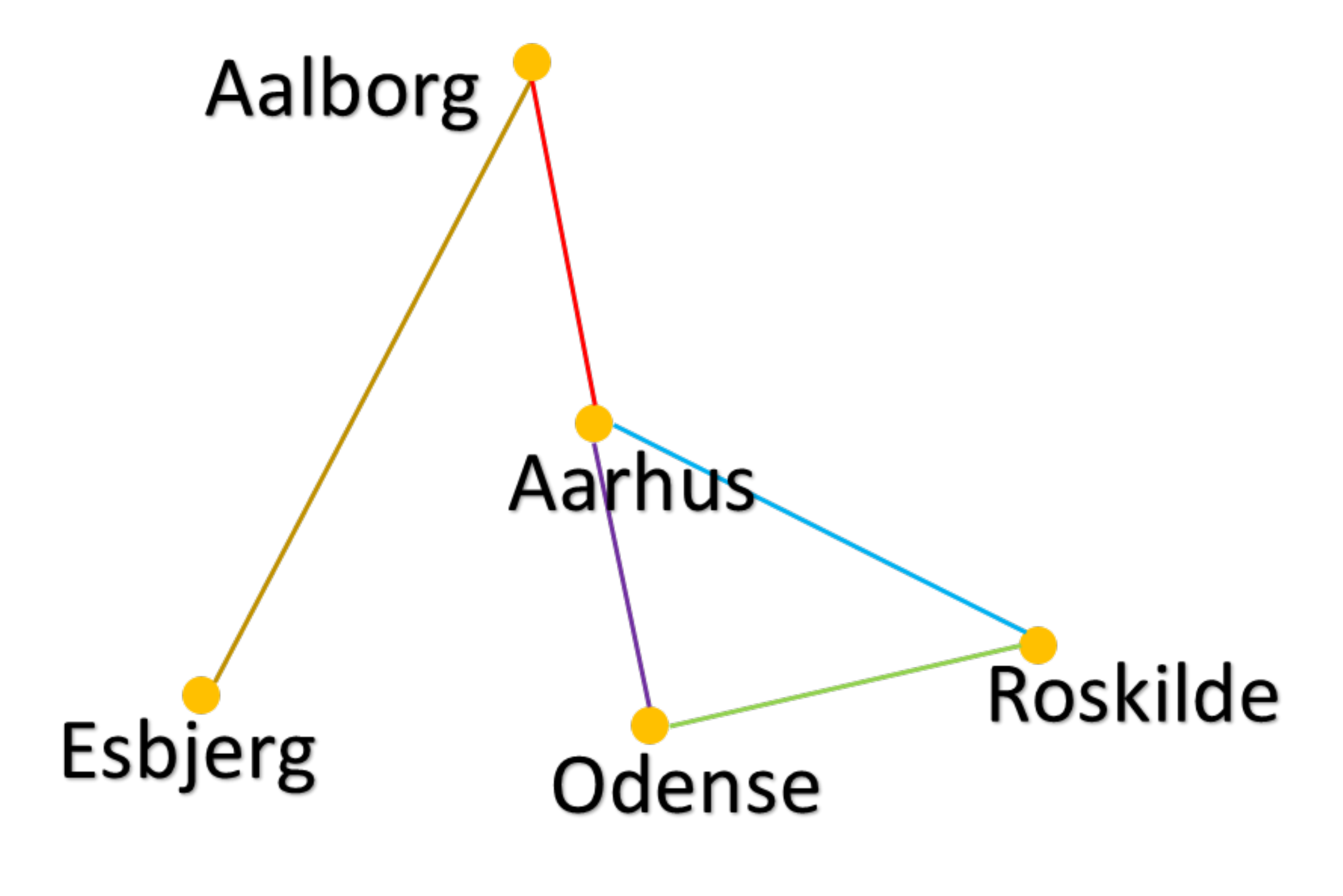}}
    \subfloat[]{\includegraphics[scale=0.33]{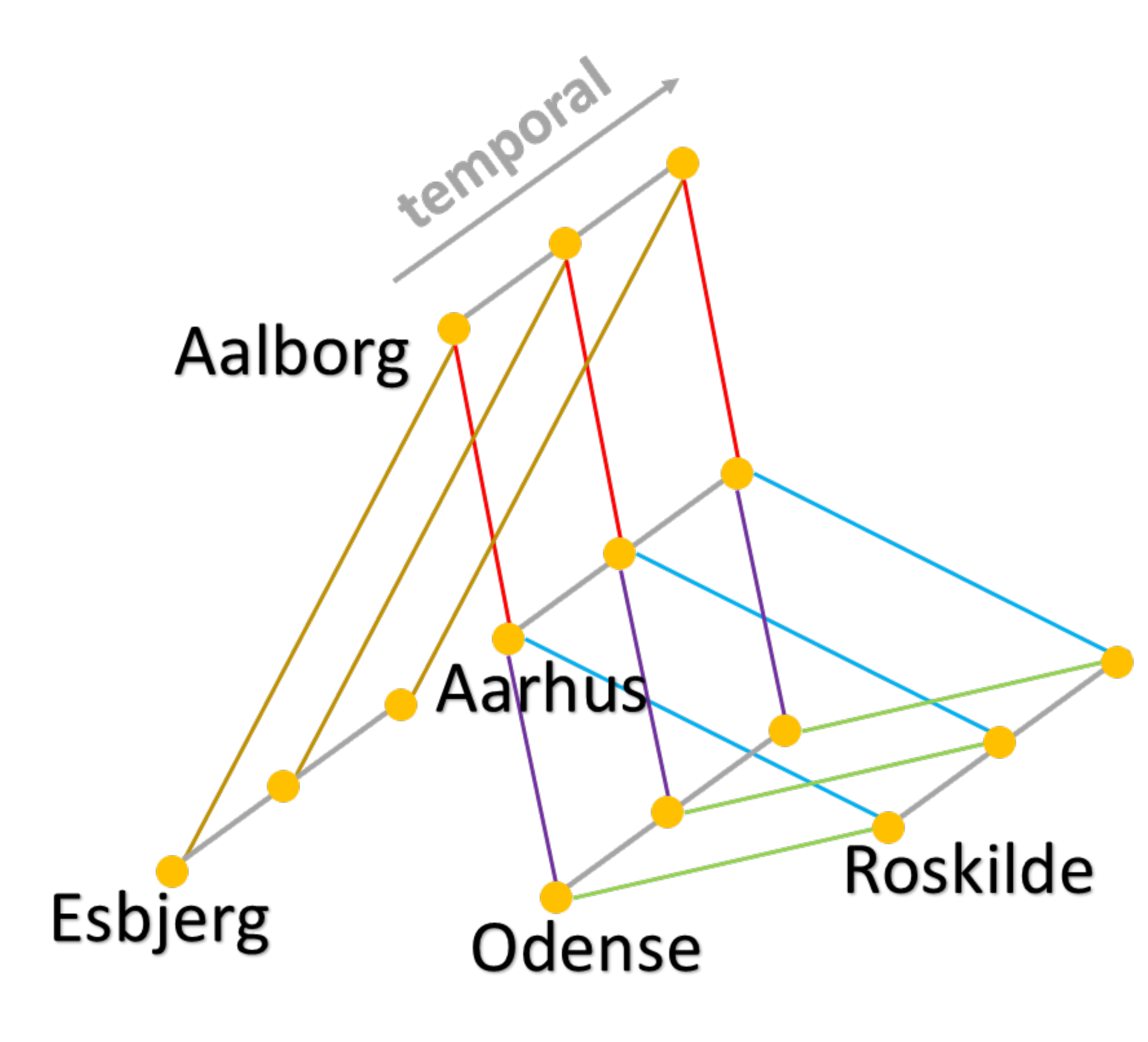}}
    \caption{Possible: (a) spatial graph; (b) spatial-temporal graph; built from Danish weather stations.}
    \label{fig:denmark_graph}
\end{figure*}

In this work we experiment on two Dutch and Danish datasets.\footnote{The datasets used within the scope of this work are described more in detail in Section \ref{section_experiments}.}
The data consists of historical observations (time steps) for several Dutch and Danish cities and several weather variables. The aim is to predict the wind speed values for selected cities some time step ahead. At a single time step, we treat the input cities as a graph where each city is a node in the graph. Node attributes are then weather variables such as temperature, wind speed and air pressure (for the first layer of the network) and features encoded by the network (in the next layers). 
For the cities shown in Fig. 
\ref{fig:map_dk_and_nl}(a)
, at a single time step, the corresponding spatial graph could be perceived as in Fig. \ref{fig:denmark_graph}(a).

In practice, time series of historical data involve multiple time steps. Therefore, %similarly as in [ref 2s-AGCN],
we include this information by expanding the previous graph into a spatial-temporal graph as shown in Fig. \ref{fig:denmark_graph}(b). Each node in the spatial graph has now its corresponding temporal neighbors, in other words, feature values for the given city at previous and/or next time step. In this way, information from multiple time steps is incorporated in the graph.

Two types of convolutions are included in our proposed models, i.e. graph convolution (spatial convolution) and temporal convolution. In what follows, we describe the two involved convolutional operations.

\subsection{Spatial Convolution}\label{subsection_spatial_conv}
Spatial convolution aggregates the information from the spatial neighbors of the graph. The graph is represented by an adjacency matrix. When binarized, the adjacency matrix determines whether there is a connection between the particular nodes in the graph. If the elements of Adjacency matrix are not binarized, then in this case they denote strength of the relation between the nodes. Following the idea from \cite{wilson@2018} and the design from \cite{shi@2018}, we make the adjacency matrix learnable. During the optimization, the network learns the graph spatial connections between the weather stations. It should be noted that since all the entries are learnt in an end-to-end fashion, the adjacency matrix is not symmetric. The learnable adjacency matrix is further transformed during the training. 
Operations similar to those from GCNs of \cite{luxburg@2007} and \cite{kipf@2017}, involving adding the self-loop connection and normalization with degree matrix are applied. To the best of our knowledge, it is the first time to apply such transformations over the adjacency matrix which is learnable.

The new, transformed matrix is created as follows. Self-loop connection in the form of an identity matrix is added to the learnable adjacency matrix:

\begin{equation}
\label{eqn:addidentitylabel}
    \hat{A} = A + I.
\end{equation}

The resulting matrix $\hat{\text{A}}$ is then scaled as follows:

\begin{equation}
    \hat{A} = \frac{\hat{A}-\hat{A}_{min}}{\hat{A}_{max}-\hat{A}_{min}}.
\end{equation}

A diagonal node degree matrix $\hat{\text{D}}$ is computed based on the normalized matrix:

\begin{equation}
    \hat{D}_{ii} = \sum_{j}\hat{A}_{ij}.
\end{equation}

Next, we apply the symmetric normalization \cite{luxburg@2007, kipf@2017} with the degree matrix:

\begin{equation}
    \hat{D}^{-\frac{1}{2}}\hat{A}\hat{D}^{-\frac{1}{2}}.
\end{equation}

The resulting transformed matrix is used for the graph convolution operation. Input data tensor $\mathcal{X}_{in}$ with the shape of $C \times T \times V$, where $C = \#channels$ (features), $T = \#time steps$ and $V = \#graph\ vertices$ (cities) is reshaped into a matrix $X_{in}$ with dimension of $CT \times V$. The graph convolution is initially performed by multiplying the reshaped input matrix with the transformed adjacency matrix:

\begin{equation}
\label{eqn:adjmatmul}
    X_{out} = X_{in}(\hat{D}^{-\frac{1}{2}}\hat{A}\hat{D}^{-\frac{1}{2}}).
\end{equation}

The adjacency matrix is the only element controlling the incorporation of the information from multiple vertices. Briefly, the learned and transformed values in the adjacency matrix entries decide how much of information coming from one weather station has to be included in the processed information of another weather station.
Next, the output matrix $X_{out}$ is reshaped back into a tensor $\mathcal{X}_{out}$ with the shape of $C \times T \times V$. Finally, simple $1 \times 1$ 2D convolution is performed in order to add linear combinations of features channel-wise and to change the number of channels. 

Additionally, we create a separate alternative model with a learnable scalar parameter, which is multiplied with the identity matrix in the transformation process:
\begin{equation}
    \hat{A} = A + \gamma I.
\end{equation}
In this manner, we let the network decide about the need and importance of the self-loop connection, which is originally imposed in Eq. \ref{eqn:addidentitylabel}. The other elements in the alternative model remain the same as for the first proposed model.

\subsection{Temporal Convolution}
Temporal convolution aggregates the information from the temporal neighbors in the graph. For each node and its features, information from the next and/or previous time step is included. The temporal convolution is implemented as a regular 2D convolution with filter of size $k \times 1$. The value of $k$ is set to 3, as this value was experimentally found to be optimal value for our models regarding their performance on the used datasets. 
The temporal convolution includes information from only one node at a time and for three ($k=3$) time steps and all the features (node attributes).

\subsection{Spatio-Temporal Blocks}
\begin{figure}[!htbp]
    \centering
    \includegraphics[scale=0.45]{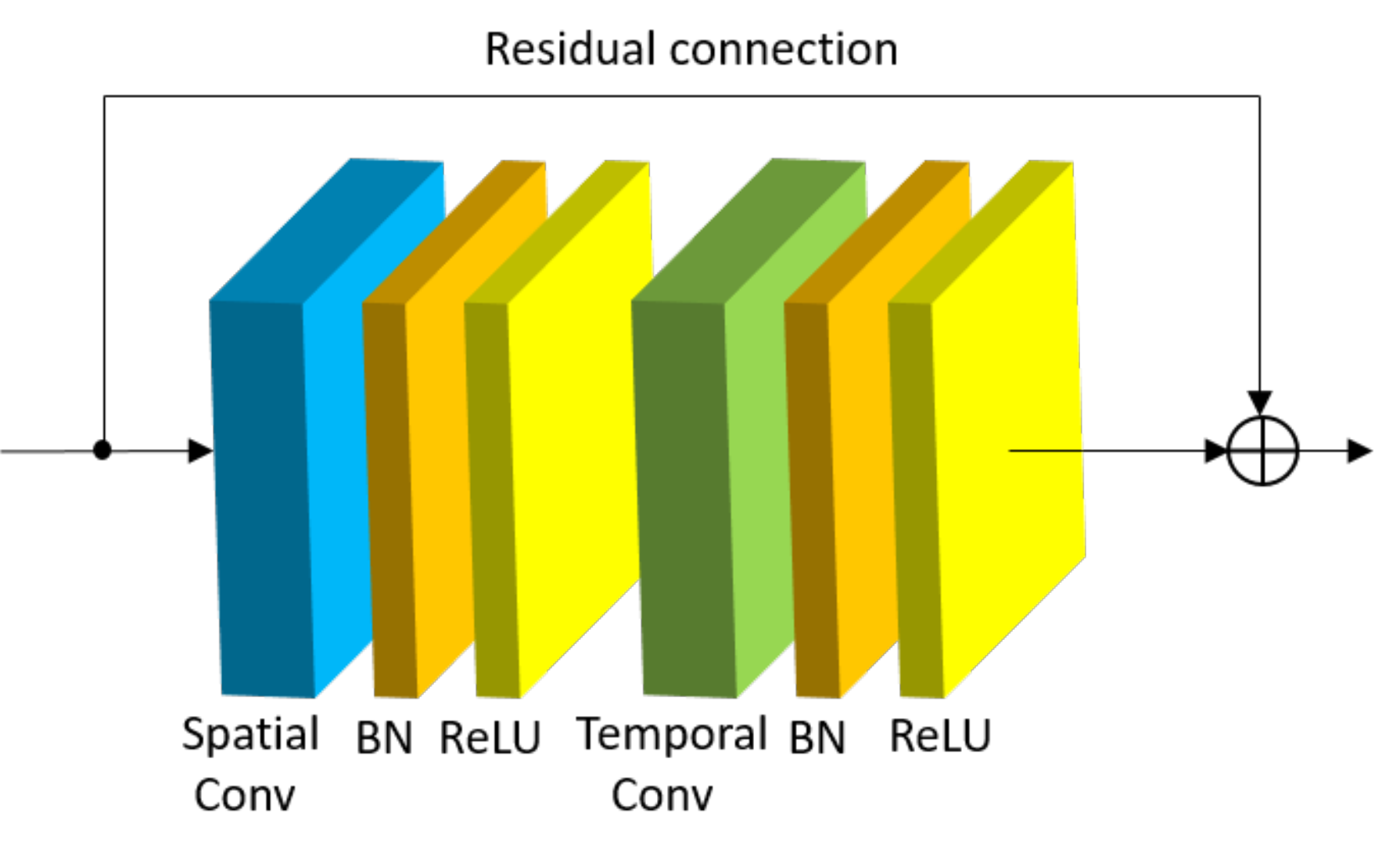}
    \caption{Spatio-temporal block (ST-block) used in our models. Figure inspired by \cite{shi@2018}.}
    \label{fig:st_block}
\end{figure}

Using spatial convolution and temporal convolution operations, we create a spatio-temporal block (ST-block). 
As presented in Fig. \ref{fig:st_block}, both spatial and temporal convolution are followed by a batch normalization (BN) layer and a ReLU activation function. Spatial convolution is performed before temporal convolution. In addition, a residual connection is added over the whole block. Each block contains its own, separate, learnable adjacency matrix within spatial convolution, as described above.

\subsection{Proposed Models}
\begin{figure*}[!htbp]
    \centering
    \includegraphics[scale=0.55]{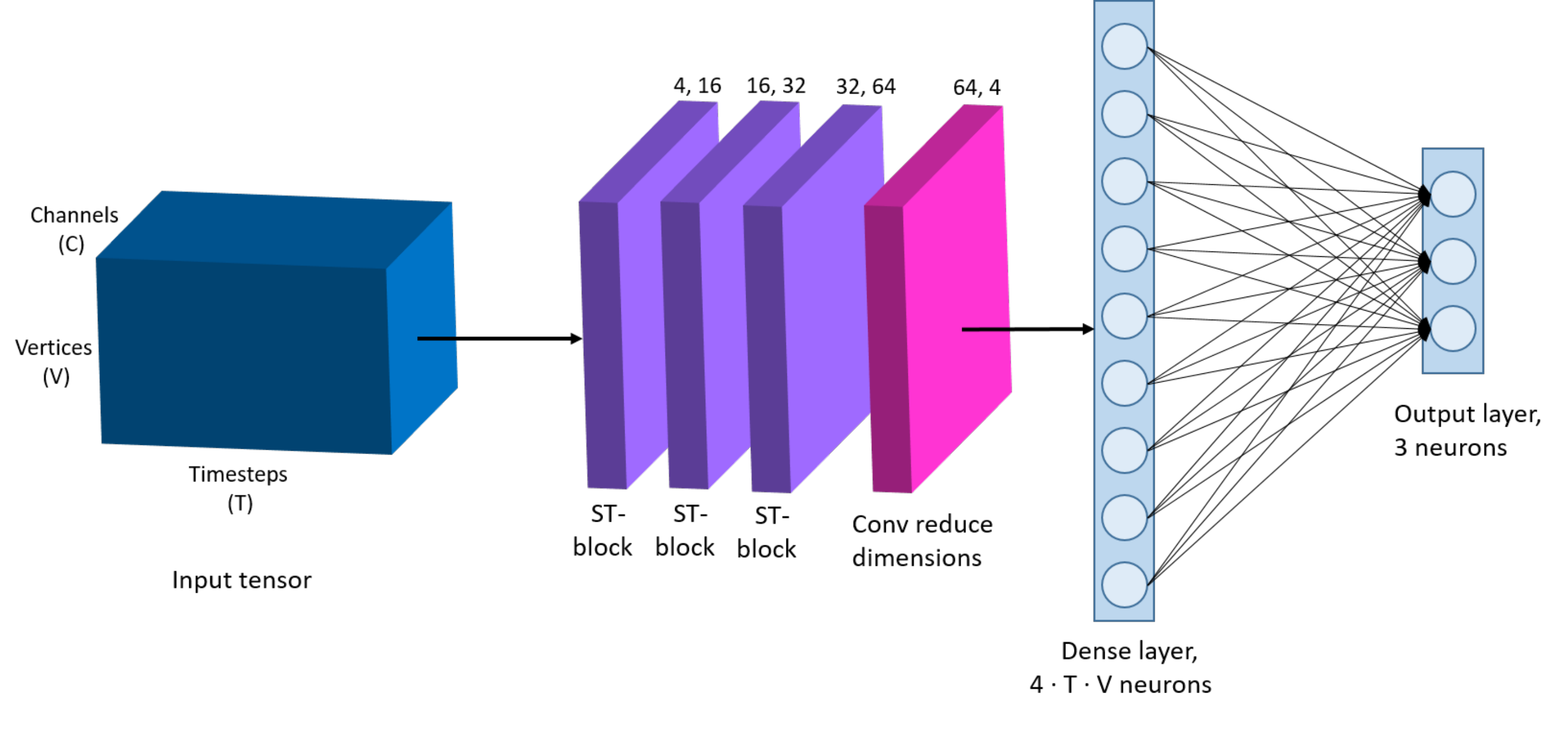}
    \caption{Overview of the proposed model(s). ST-block stands for the spatio-temporal block. The numbers above the blocks indicate the number of input and output channels respectively. The number of output neurons is specific for the Danish cities dataset (3 output cities).}
    \label{fig:proposed_model}
\end{figure*}

We stack three spatio-temporal blocks with the following number of output channels: 16, 32 and 64. The last block is followed by $1 \times 1$ 2D convolution reducing the number of channels to 4 before flattening. A fully connected (FC) layer is added at the end of the network to obtain wind speed predictions for for selected output cities. The model architecture is presented in Fig. \ref{fig:proposed_model}. We call our first proposed model \textbf{WeatherGCNet}. 
As stated in Section \ref{subsection_spatial_conv}, we also create an alternative model with an additional learnable parameter $\gamma$ inside the spatial convolution block. This model is refered to as \textbf{WeatherGCNet with $\gamma$}.
The following setup is applied for the training procedure. Batch size is set to 64. Adam optimizer is used with the default value of learning rate set to 0.001. 
The number of input timesteps $T$ processed by our models is set to 30. 
The details of the used datasets as well as the evaluation are described in the next section. The implementation of our proposed architecture and the trained models are available on Github.\footnote{\url{https://github.com/tstanczyk95/WeatherGCNet}}

\section{Experiments}\label{section_experiments}
We consider two datasets for the task of wind speed prediction. The data comes from weather stations located in Denmark and the Netherlands.

The Danish dataset contains hourly measurements with the values of the following weather variables: temperature, air pressure, wind speed and wind direction. Each measurement includes five cities in Denmark, as presented in Fig. \ref{fig:map_dk_and_nl}(a). The data covers time span of years 2000-2010. The data from 2000-2009 are used for the training and validation set and the data from 2010 is used for the test set. The dataset is normalized using min-max normalization based on values coming from the training set. One sample of the dataset contains values of all aforementioned weather variables, for all cities and for user-defined-number of historical time steps. In this study, the target is set to wind speed in three cities: Esbjerg, Odense, Roskilde. The input is the tensor with the shape of $C \times T \times V$, where $C = \#channels$ (weather variables), $T = \#time steps$, $V = \#cities$ (treated as vertices in the graph). 
This dataset is available online.\footnote{\url{https://sites.google.com/view/siamak-mehrkanoon/code-data}}

The Dutch dataset contains hourly measurements of the following weather variables: wind speed, wind direction, temperature, dew point, air pressure and rain amount. Each measurement include seven cities in the Netherlands as presented in Fig. \ref{fig:map_dk_and_nl}(b). The data covers the time period between January 1, 2011 and March 29, 2020. Samples from January 2011 to December 2018 are used for the training and validation set, and samples from January 2019 to March 2020 are used for the test set. The data is normalized using min-max normalization before it is fed to the network. 
The input data arrangement is the same as for the Danish weather stations, with the shape of $C \times T \times V$. The target is wind speed for all weather stations. This dataset is available online.\footnote{\url{https://github.com/HansBambel/multidim_conv}}

We evaluate our models together with the other models presented in \cite{trebing2020wind}. The referenced models are trained and evaluated on the same datasets using the same data splits as used for our models. Analogously as in \cite{trebing2020wind}, we report mean absolute error (MAE) and mean squared error (MSE) for each model on the test set. 
The reported MAEs and MSEs are obtained by taking the average over all output cities. The MAE and MSE are defined as follows:

\begin{equation}
   MAE = \frac{\sum_{i=1}^{n} |y_{i} - \hat{y_{i}}|}{n},
\end{equation}
\begin{equation}
   MSE = \frac{\sum_{i=1}^{n} (y_{i} - \hat{y_{i}})^{2}}{n}.
\end{equation}
Here, $n$ denotes the number of samples in the test set. $y_{i}$ and $\hat{y_{i}}$ denote ground truth and predicted value respectively. For the Danish dataset, we train the models for predicting wind speed over 6, 12, 18 and 24 hours ahead. The obtained MAE and MSE on the test set for wind speed are tabulated in Table \ref{tab:results_dk_wind}. The best MAE and MSE are underlined for each prediction time. 
The scores of the baseline models are taken from the original paper \cite{trebing2020wind}.

\begin{table*}[!htbp]
    \centering
    \scriptsize{
    \renewcommand{\arraystretch}{1.5}
    \caption{The average MAEs and MSEs of evaluated models for wind speed prediction over the Danish cities dataset.}
    \label{tab:results_dk_wind}
    %\resizebox{\columnwidth}{!}{
    \begin{tabular}{l | l l l l | l l l l}
    \Xhline{3\arrayrulewidth}
    &\multicolumn{4}{c|}{\textbf{MAE}}
    &\multicolumn{4}{c}{\textbf{MSE}} \\
    \Xhline{3\arrayrulewidth}
    \multirow{2}{*}{\textbf{Model}}& 
    \multirow{2}{*}{\textbf{6h ahead}} & 
    \multirow{2}{*}{\textbf{12h ahead}} & 
    \multirow{2}{*}{\textbf{18h ahead}} & 
    \multirow{2}{*}{\textbf{24h ahead}} &
    \multirow{2}{*}{\textbf{6h ahead}} & 
    \multirow{2}{*}{\textbf{12h ahead}} & 
    \multirow{2}{*}{\textbf{18h ahead}} & 
    \multirow{2}{*}{\textbf{24h ahead}} \\
     & & & & \\\Xhline{3\arrayrulewidth}
         \textbf{2D} & 1.304 & 1.746 & 1.930 & 2.004 & 2.824 & 5.088 & 6.120 & 6.610\\ 
         \textbf{2D + Attention} & 1.313 & 1.715 & 1.905 & 1.950 & 2.885 & 4.896 & 5.933 & 6.201 \\  
         \textbf{2D + Upscaling} & 1.307 & 1.723 & 1.858 & 1.985 & 2.826 & 4.931 & 5.639 & 6.474 \\  
         \textbf{3D} & 1.311 & 1.677 & 1.908 & 1.957 & 2.855 & 4.595 & 5.958 & 6.238 \\
         \textbf{Multidimensional} & 1.302 & 1.706 & 1.873 & 1.925 & 2.804 & 4.779 & 5.773 & 6.066 \\
         \textbf{WeatherGCNet} & 1.279 & 1.638 & 1.777 & 1.869  & 2.698 & 4.407 & 5.148 & 5.641 \\
         \textbf{WeatherGCNet with $\gamma$} & \underline{1.267} & \underline{1.616} & \underline{1.767} & \underline{1.853} & \underline{2.684} & \underline{4.285} & \underline{5.096} & \underline{5.566} \\
        \Xhline{3\arrayrulewidth}
    \end{tabular}
    }
   % }
\end{table*}

\begin{table*}[!htbp]
    \centering
    \scriptsize{
    \renewcommand{\arraystretch}{1.5}
    \caption{The average MAEs and MSEs of evaluated models for wind speed prediction over the Dutch cities dataset.}
    \label{tab:results_nl_wind}
    %\resizebox{\columnwidth}{!}{
    \begin{tabular}{l | l l l l l | l l l l l}
    \Xhline{3\arrayrulewidth}
    &\multicolumn{5}{c|}{\textbf{MAE}}
    &\multicolumn{5}{c}{\textbf{MSE}} \\
    \Xhline{3\arrayrulewidth}
    \multirow{2}{*}{\textbf{Model}}& 
    \multirow{2}{*}{\textbf{2h ahead}} & 
    \multirow{2}{*}{\textbf{4h ahead}} & 
    \multirow{2}{*}{\textbf{6h ahead}} & 
    \multirow{2}{*}{\textbf{8h ahead}} &
    \multirow{2}{*}{\textbf{10h ahead}} & 
    \multirow{2}{*}{\textbf{2h ahead}} & 
    \multirow{2}{*}{\textbf{4h ahead}} &
    \multirow{2}{*}{\textbf{6h ahead}} &
    \multirow{2}{*}{\textbf{8h ahead}} & 
    \multirow{2}{*}{\textbf{10h ahead}} \\
     & & & & & \\\Xhline{3\arrayrulewidth}
         \textbf{2D} & 8.18 &	10.08 &	12.03 &	13.15 &	14.51 & 118.35 &	179.38 &	253.60 &	303.28 &	369.16\\
         \textbf{2D + Attention} & 8.10 &	10.09 &	11.83 &	13.10 &	14.13 & 116.09 &	180.69 &	247.22 &	300.41 &	351.26\\
         \textbf{2D + Upscaling} & 8.24 &	10.22 &	11.83 &	13.74 &	14.80 & 120.65 &	183.87 &	248.85 &	332.87 &	387.92\\
         \textbf{3D} & 8.05 &	10.15 &	11.93 &	13.01 &	14.24 & 115.17 &	183.55 &	251.11 &	294.39 &	355.44\\
         \textbf{Multidimensional} & 8.10 &	10.03 &	11.46 &	12.79 &	13.81 & 115.92 &	178.39 &	228.97 &	283.13 &	336.29\\
         \textbf{WeatherGCNet} & \underline{7.96} &	9.97 &	11.16 &	\underline{12.30} &	\underline{13.33} & \underline{111.83} &	174.55 &	219.45 &	\underline{265.71} &	\underline{309.82}\\
         \textbf{WeatherGCNet with $\gamma$}  & 7.97 &	\underline{9.74} &	\underline{10.99} &	12.44 &	13.55 & 113.32 &	\underline{168.08} &	\underline{212.30} &	272.92 &	319.05\\
        \Xhline{3\arrayrulewidth}
    \end{tabular}
    }
   % }
\end{table*}

It can be observed that the proposed models, WeatherGCNet and WeatherGCNet with $\gamma$, outperform all the other baseline models in both MAE and MSE scores. This is due to the fact that our models learn and incorporate the spatial neighbor relations between the weather stations based on a graph form rather than tensorial form.

For the Dutch dataset, we train the models for predicting wind speed over 2, 4, 6, 8 and 10 hours ahead. The obtained MAE and MSE scores for the prediction of wind speed  are tabulated in Table \ref{tab:results_nl_wind}. The relevant scores of the baseline models are initially unavailable. We use the authors’ publicly available code implementation to perform appropriate training and collect the corresponding MAE and MSE scores.

Similarly as for the Danish dataset, it can be seen that both, WeatherGCNet and WeatherGCNet with $\gamma$, outperform the other baselines for the Dutch dataset. 

\begin{figure*}[!htbp]
\centering
    \subfloat[]{\includegraphics[scale=0.3]{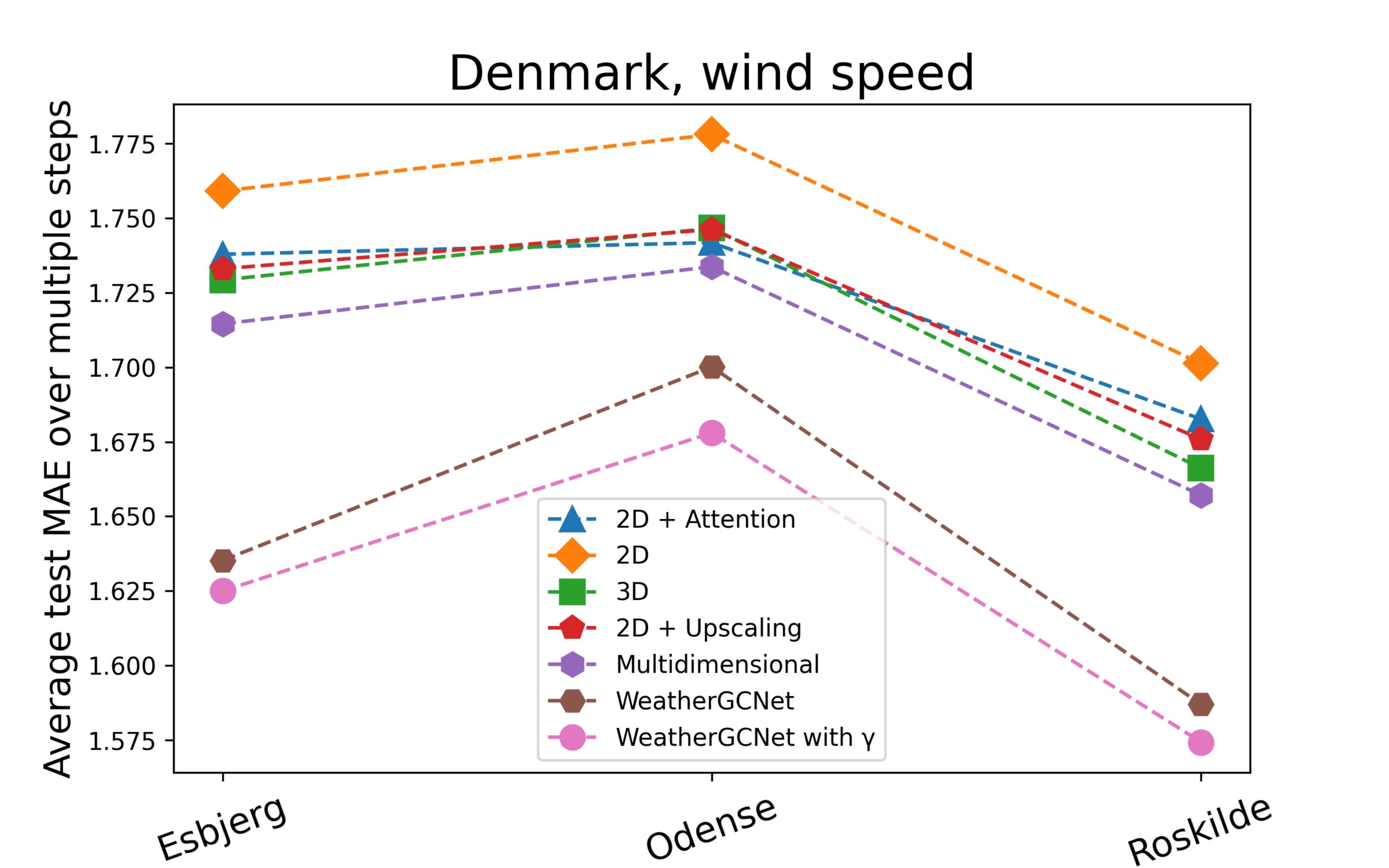}}
    \subfloat[]{\includegraphics[scale=0.3]{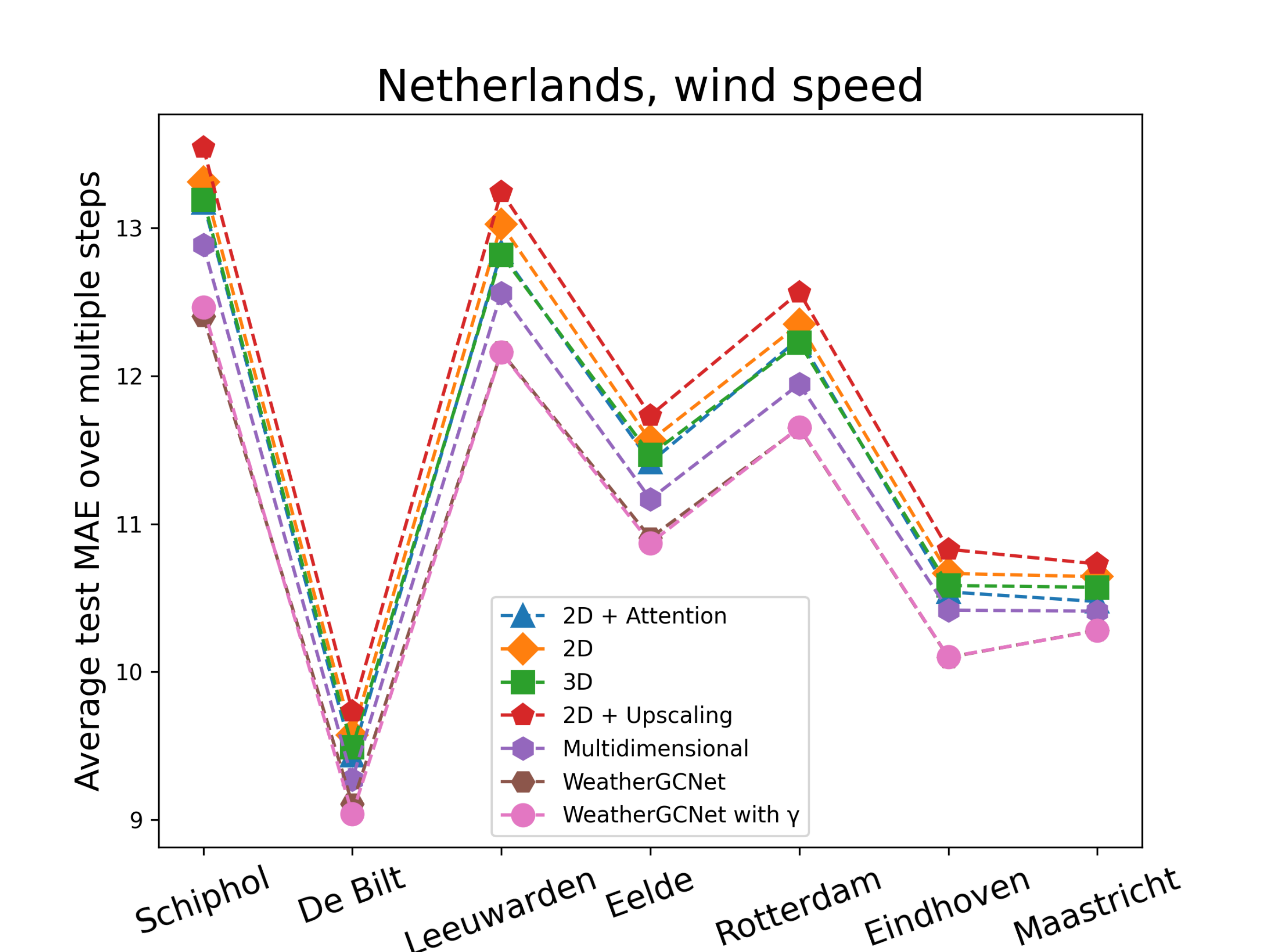}}
    \\\vskip 0.5pt plus 0.25fil
        \caption{Test MAE of the models per city averaged over all considered prediction time steps.}
    \label{fig:average_errors_per_city}
\end{figure*}

Fig. \ref{fig:average_errors_per_city} shows the performance of the baseline models and our proposed models per weather station. The scores (MAE errors) are averaged over all considered prediction time steps. From Fig.  \ref{fig:average_errors_per_city}, one can observe that our proposed models, WeatherGCNet and WeatherGCNet with $\gamma$ perform better than the baselines for each city.

\begin{figure*}[!htbp]
\centering
    \subfloat[]{\includegraphics[scale=0.55]{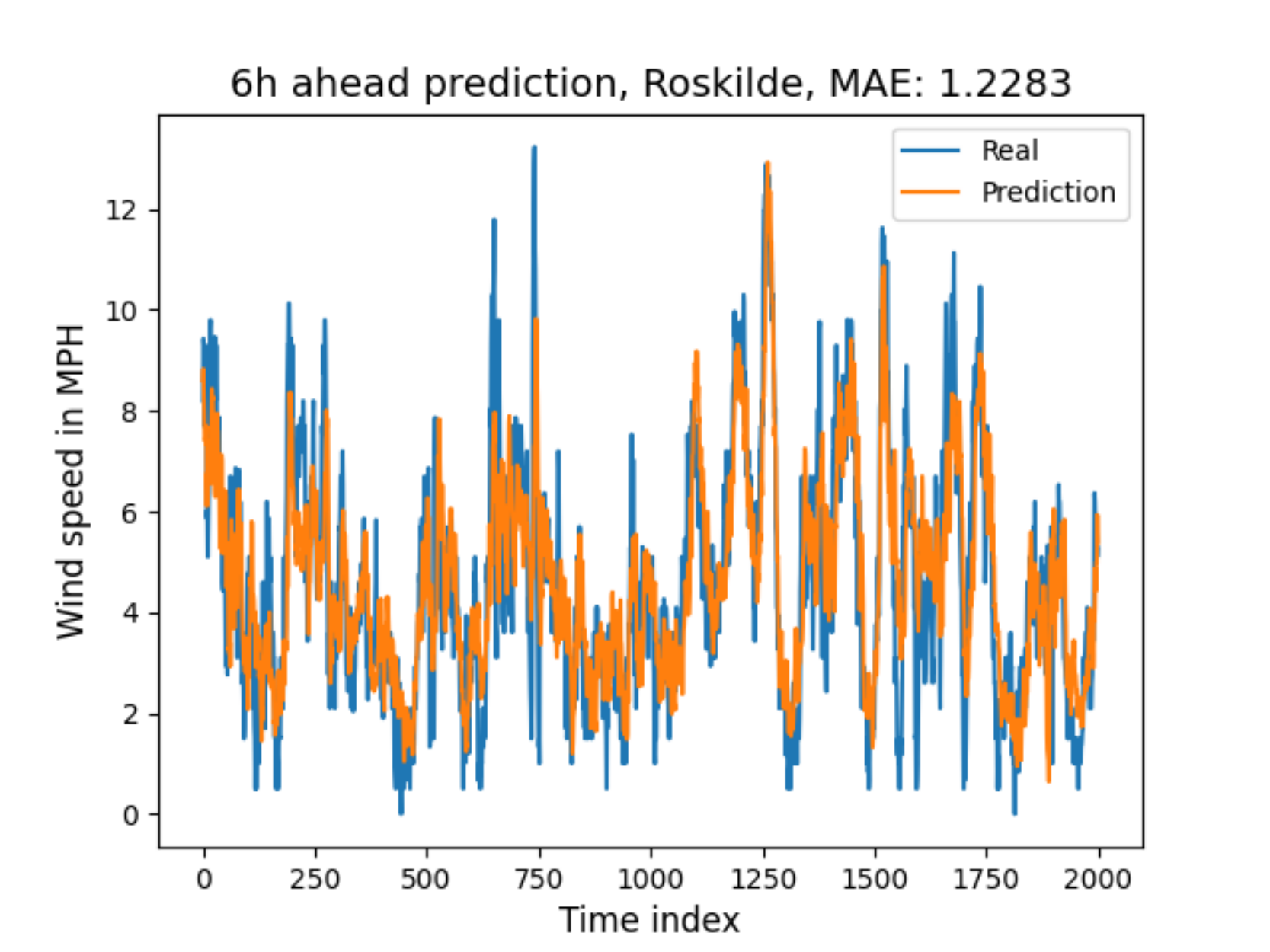}}
    % \hspace{0.2cm}
    \subfloat[]{\includegraphics[scale=0.55]{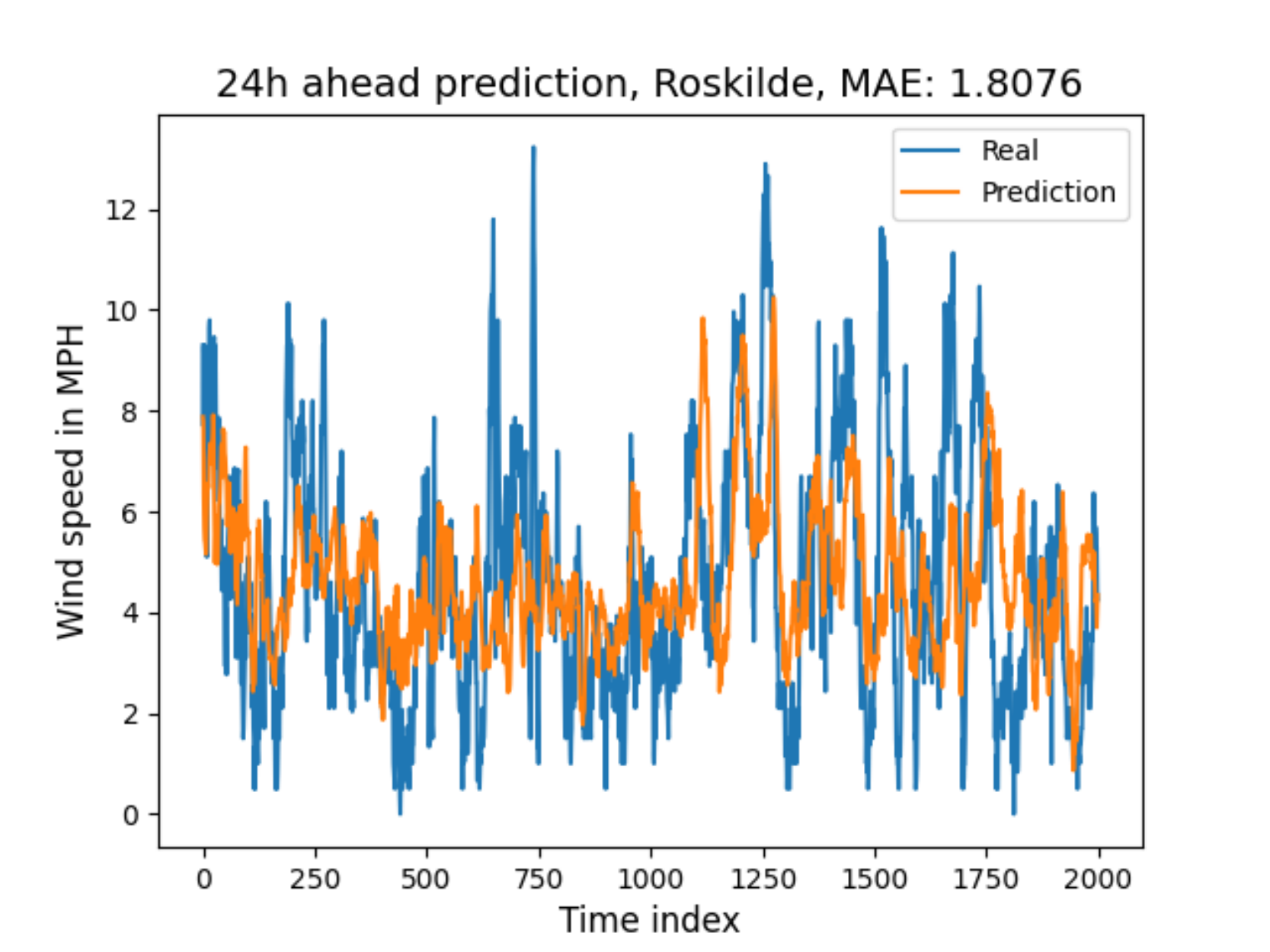}}
    \\\vskip 0.5pt plus 0.25fil
    \caption{WeatherGCNet (without $\gamma$) wind speed predictions for Roskilde (DK) for: (a) 6h ahead, (b) 24h ahead.}
    \label{fig:dk_plots}
\end{figure*}

\begin{figure*}[!htbp]
\centering
    \subfloat[]{\includegraphics[scale=0.55]{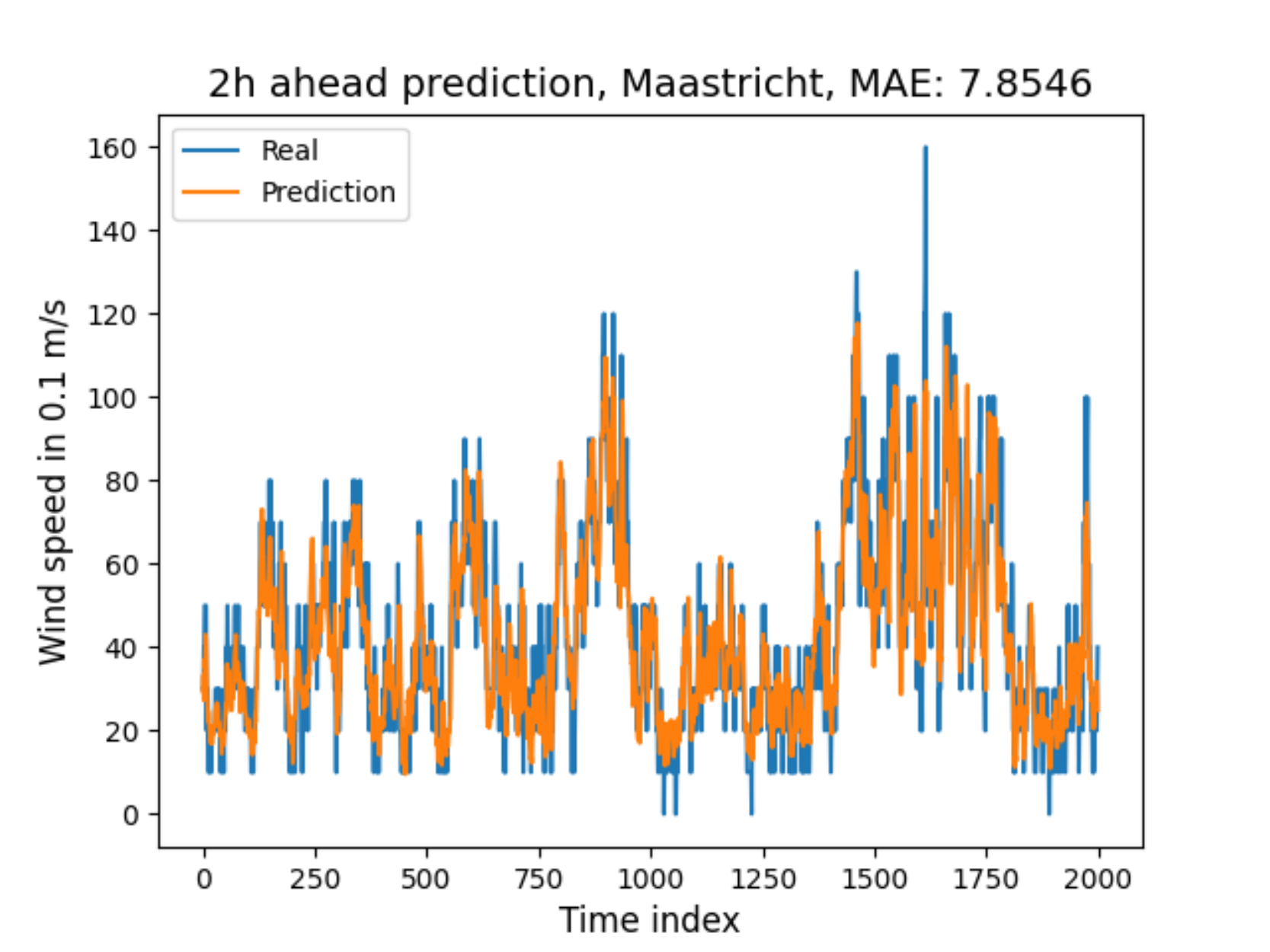}}
    % \hspace{0.2cm}
    \subfloat[]{\includegraphics[scale=0.55]{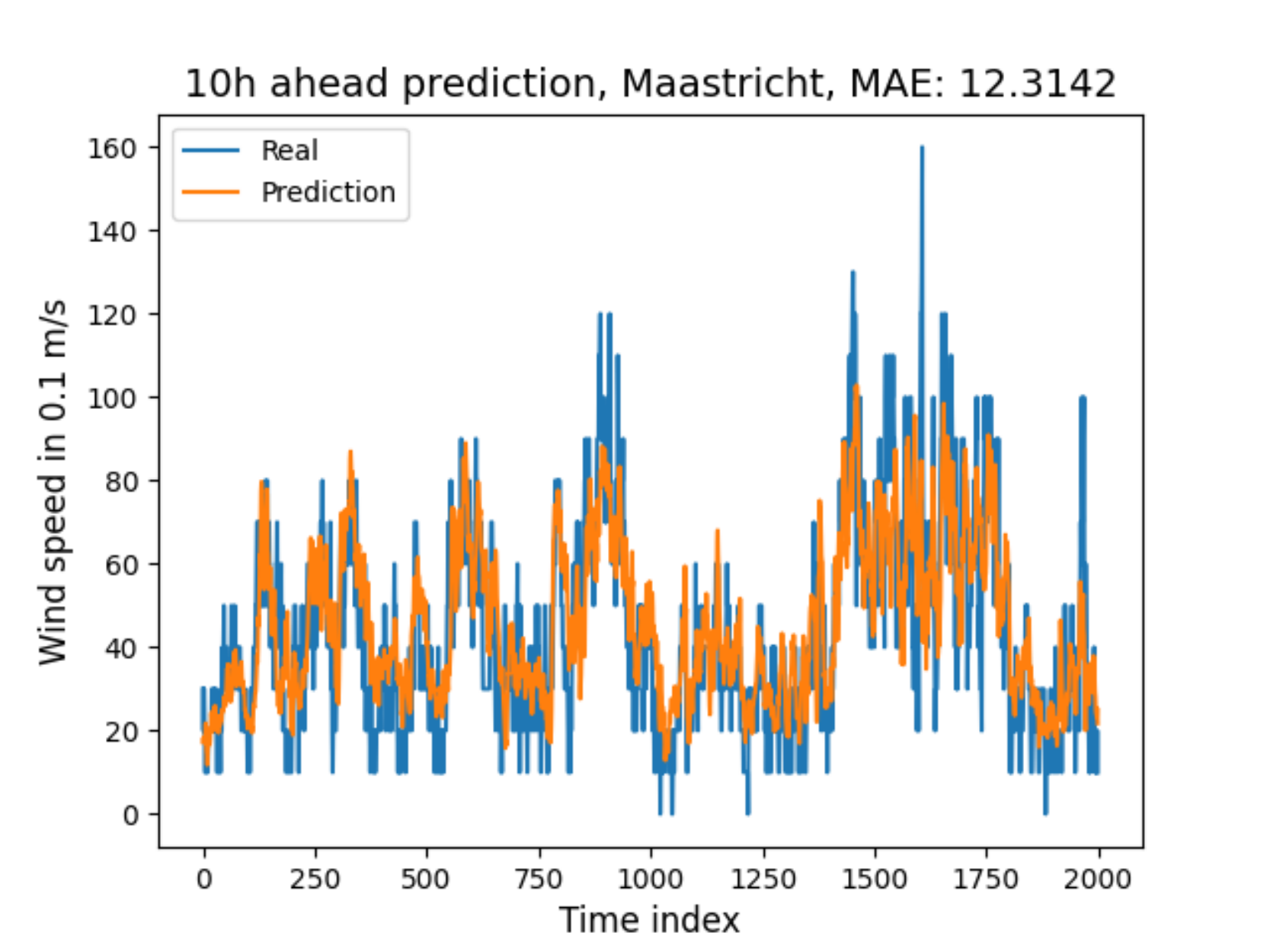}}
    \\\vskip 0.5pt plus 0.25fil
    \caption{WeatherGCNet (without $\gamma$) wind speed predictions for Maastricht (DK) for: (a) 2h ahead, (b) 10h ahead.}
    \label{fig:nl_plots}
\end{figure*}

Fig. \ref{fig:dk_plots}(a) and \ref{fig:dk_plots}(b) show plots of wind speed prediction obtained by the proposed model for the selected Danish City Roskilde. Fig \ref{fig:nl_plots}(a) and \ref{fig:nl_plots}(b) show wind speed prediction for selected Dutch city Maastricht. For all the above-mentioned figures, WeatherGCNet (without $\gamma$) is used. It should be noted that a subset of test data is used for plotting whereas MAE scores included in the figures are reported for the whole test set.
The MAE scores indicate average absolute differences between model predictions and corresponding ground truth values in the included units for expressing wind speed values (see the vertical axes of Fig. \ref{fig:dk_plots} and \ref{fig:nl_plots}). E.g. the MAE score of $7.8546$ presented in \ref{fig:nl_plots}(a) means that on the average (all samples in the test set) absolute difference between the model predictions and ground truth values is $7.8546$ of $0.1 m/s$ and thus it is equivalent to $0.78546 m/s$.

\section{Discussion}\label{section_discussion}
The proposed graph convolutional architectures outperform 
the referred baselines for the studied datasets in all cases examined.
The obtained results show that by treating the weather stations as the nodes in a graph, the proposed models better learn the underlying spatial relations between the weather stations which result in improving the prediction accuracy. In the case of referenced baseline models \cite{trebing2020wind}, the convolution operations are performed over the input tensor. Neighborhood and relation between the weather stations are determined by the order of the items in the dataset. 
% we keep this if they asked for further explaination:
%If the weather station in the Dutch cities dataset are included in the following order: Schiphol, De Bilt, Leeuwarden, Eelde, Rotterdam, Eindhoven, Maastricht, and the kernel size along the city dimension is set to 3, then each city has (at most) 2 neighbors – the adjacent cities in the dataset. E.g., for Leeuwarden, the only neighboring cities are De Bilt and Eelde. In case of Eelde, the neighboring cities are Leeuwarden and Rotterdam, etc. The entries of kernel used for the convolution have different values (yet it is used for all cities). However, the neighborhood is determined simply by the order of the cities in the dataset and the strength of relation between the neighbors is always the same, and it can be perceived as a weight with value of 1.0.
On the other hand, with the proposed graph based approach, one can use an adjacency matrix to define the relations between the cities, the number of neighbors and the strength of each relation. In our proposed models we make the adjacency matrix learnable, so that the network can decide about the strength of the relations on its own based on the historical weather data observations. From Tables \ref{tab:results_dk_wind}-\ref{tab:results_nl_wind}, one can observe that the approach with learnable adjacency matrix indicates the superiority of the proposed models over the previous baseline models. 

\begin{figure*}[!htbp]
\centering
    \subfloat[]{\includegraphics[scale=0.53]{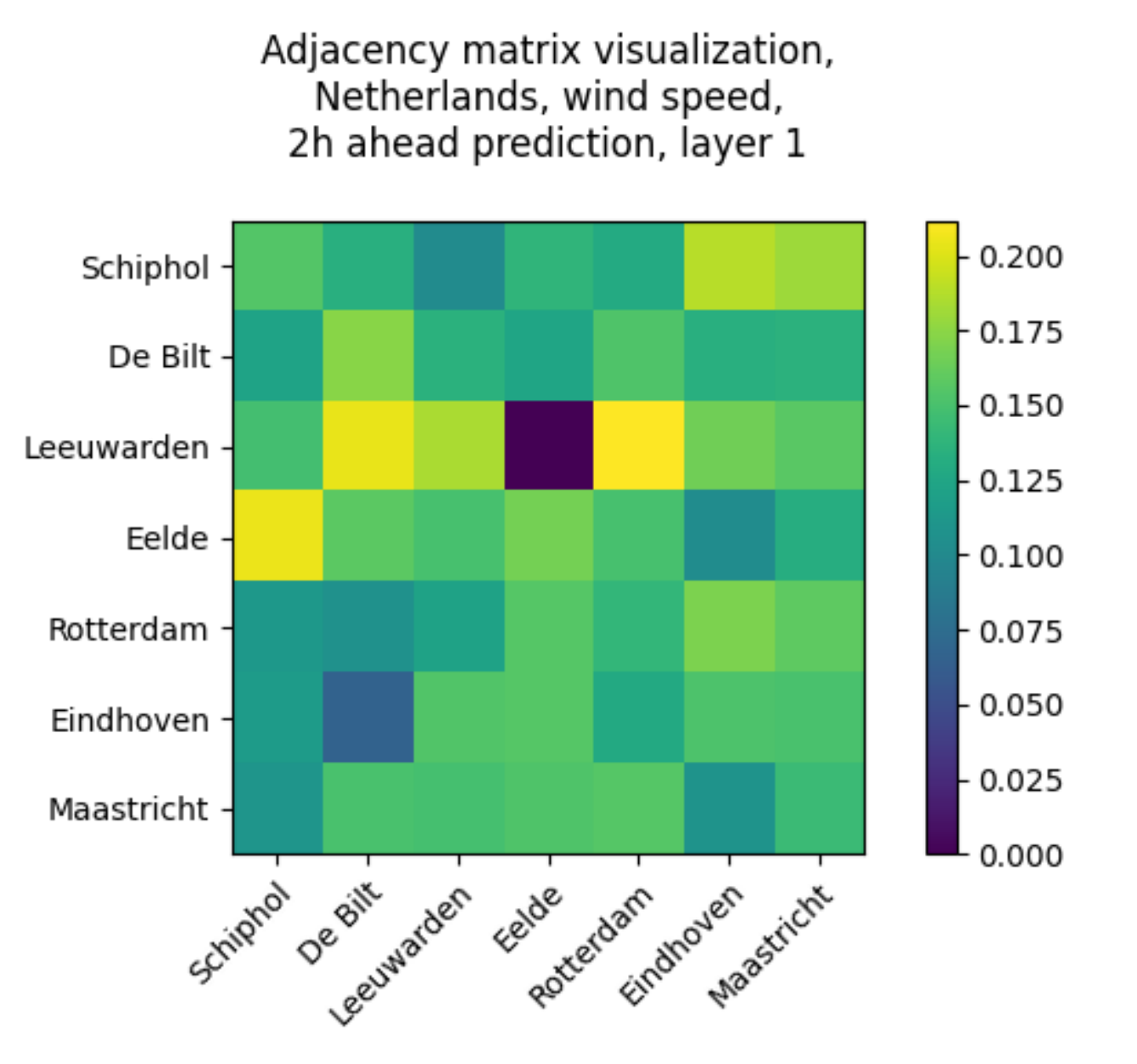}}
    \hspace{0.1cm}
    \subfloat[]{\includegraphics[scale=0.53]{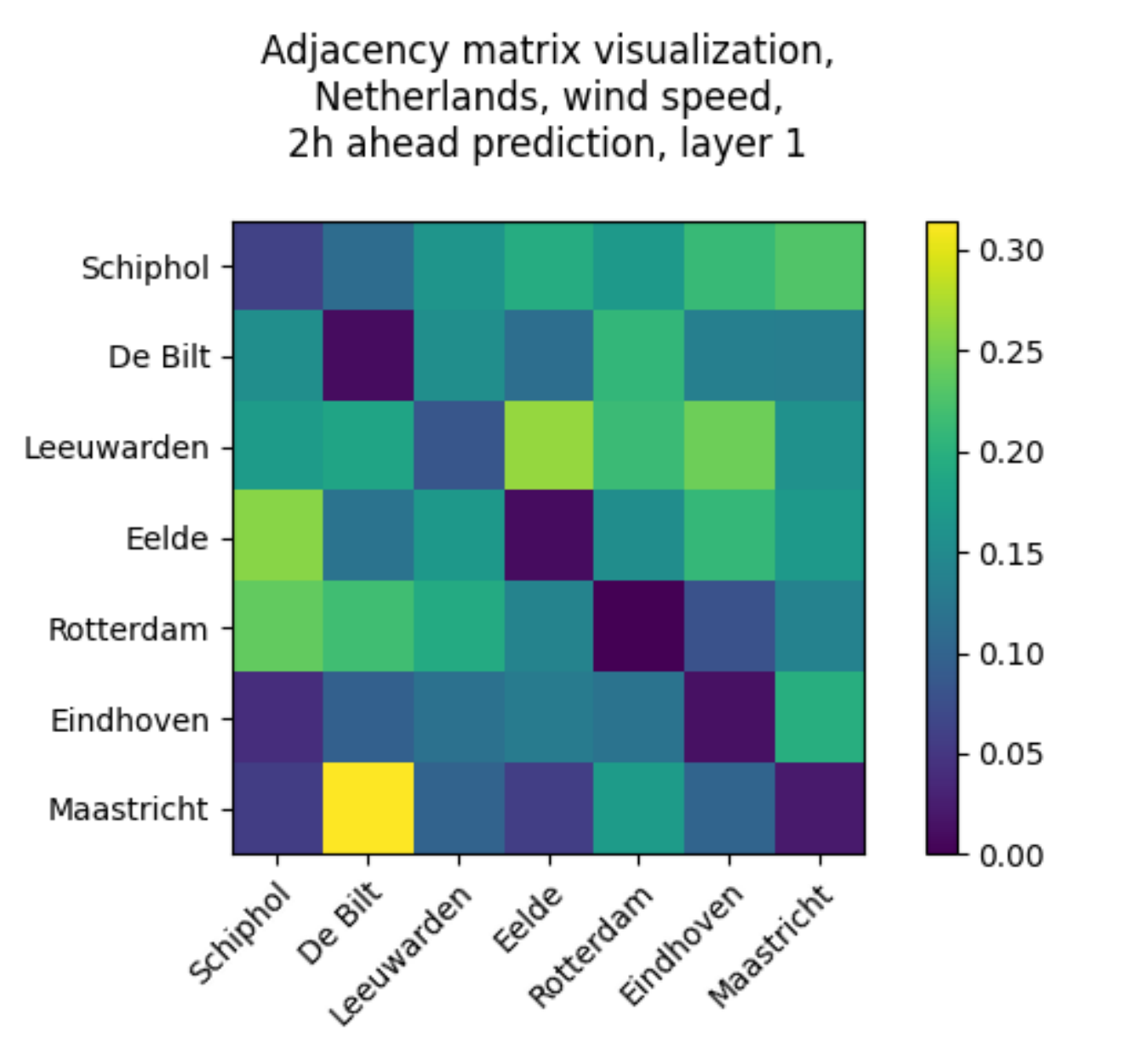}}
    \\\vskip 0.5pt plus 0.25fil
    \subfloat[]{\includegraphics[scale=0.53]{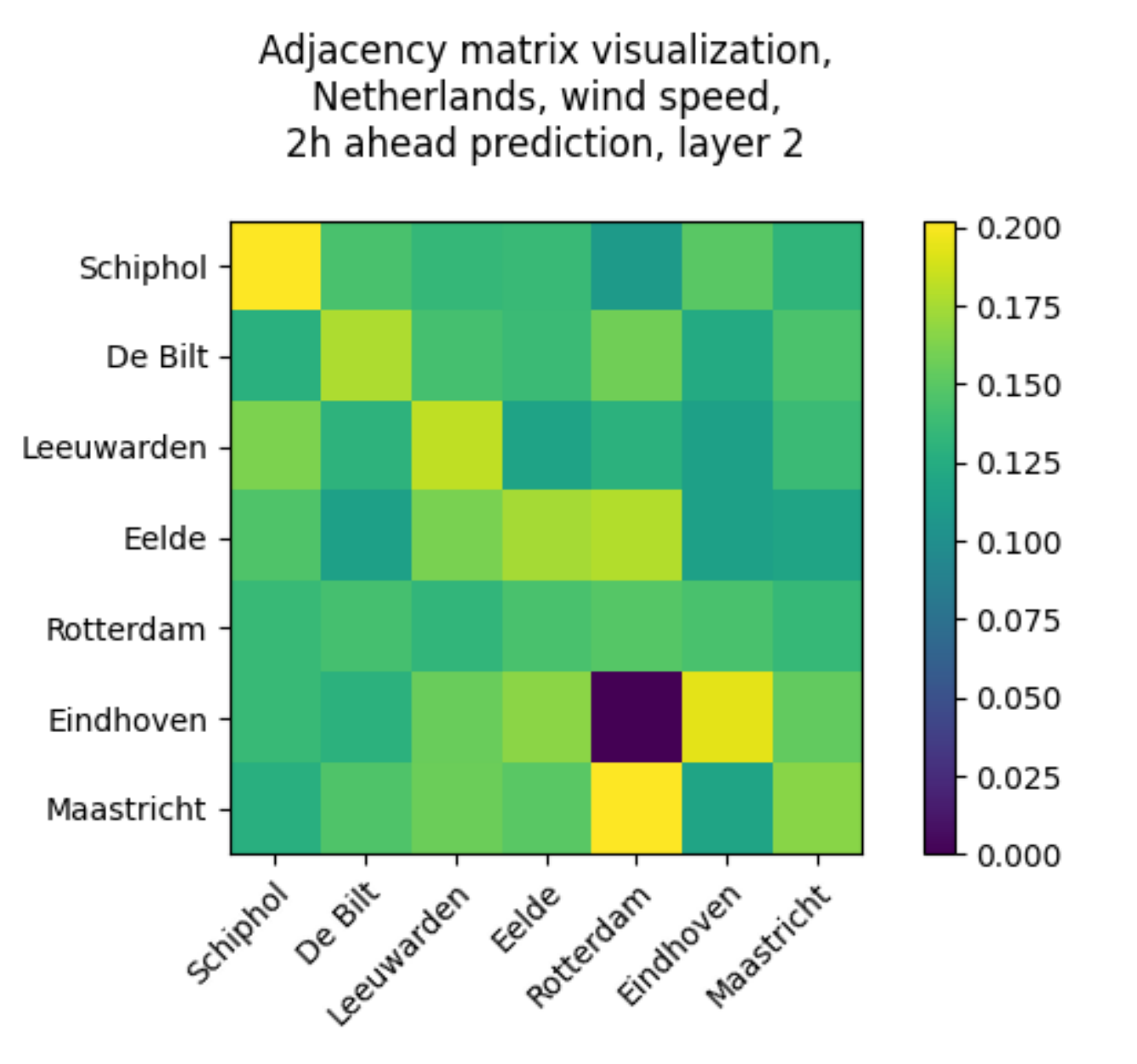}}
    \hspace{0.1cm}
    \subfloat[]{\includegraphics[scale=0.53]{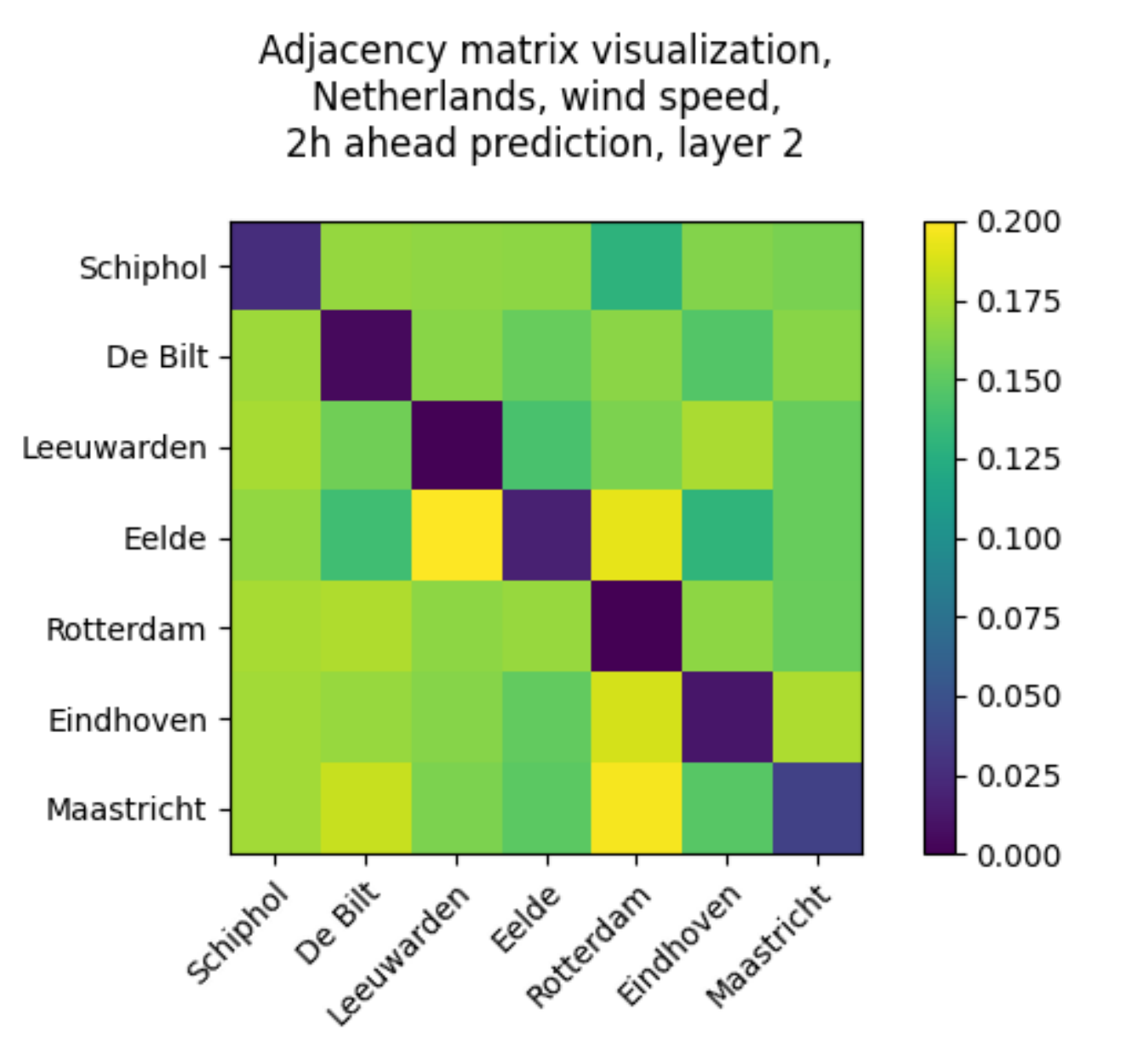}}
    \\\vskip 0.5pt plus 0.25fil
    \subfloat[]{\includegraphics[scale=0.53]{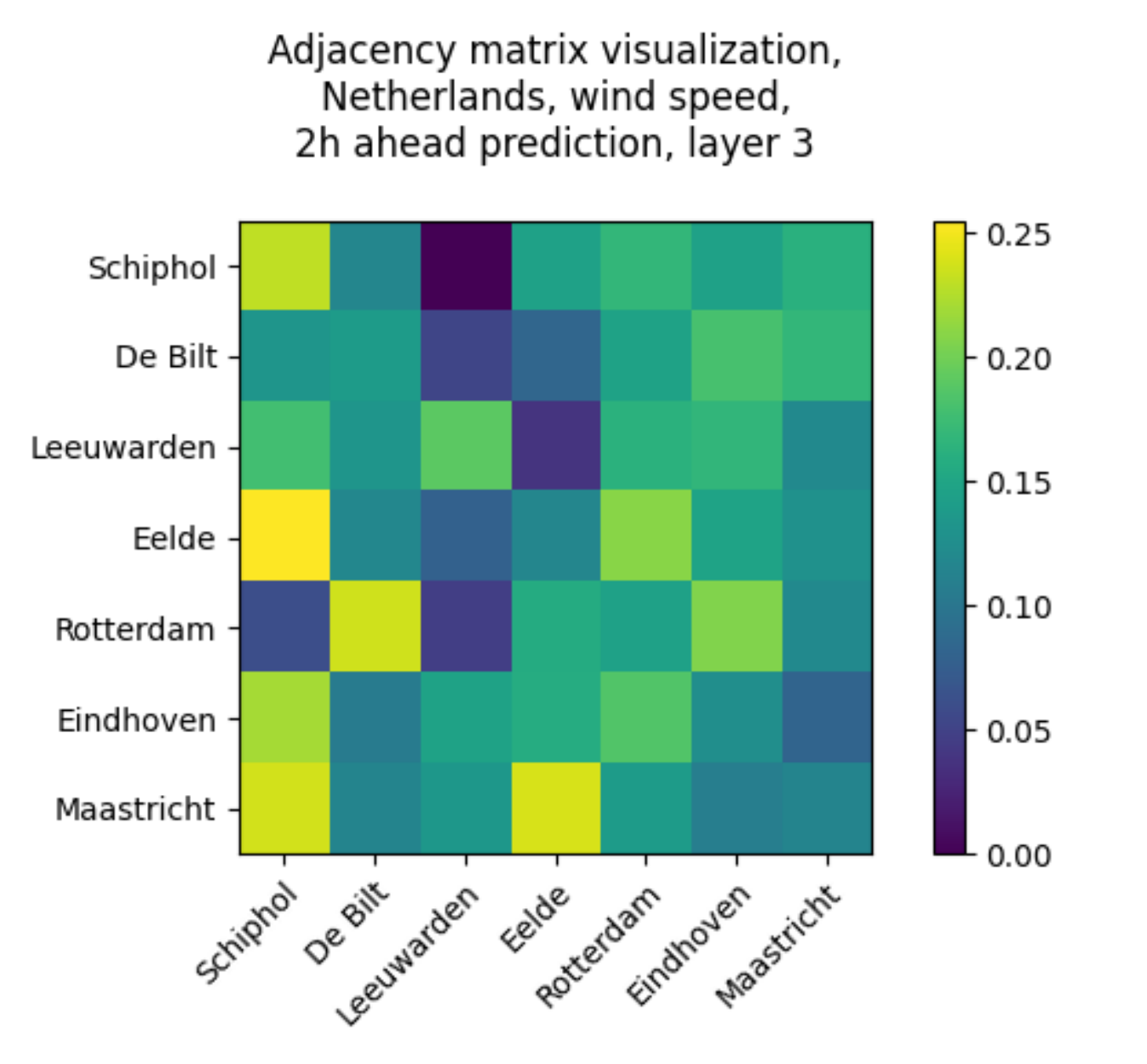}}
    \hspace{0.1cm}
    \subfloat[]{\includegraphics[scale=0.53]{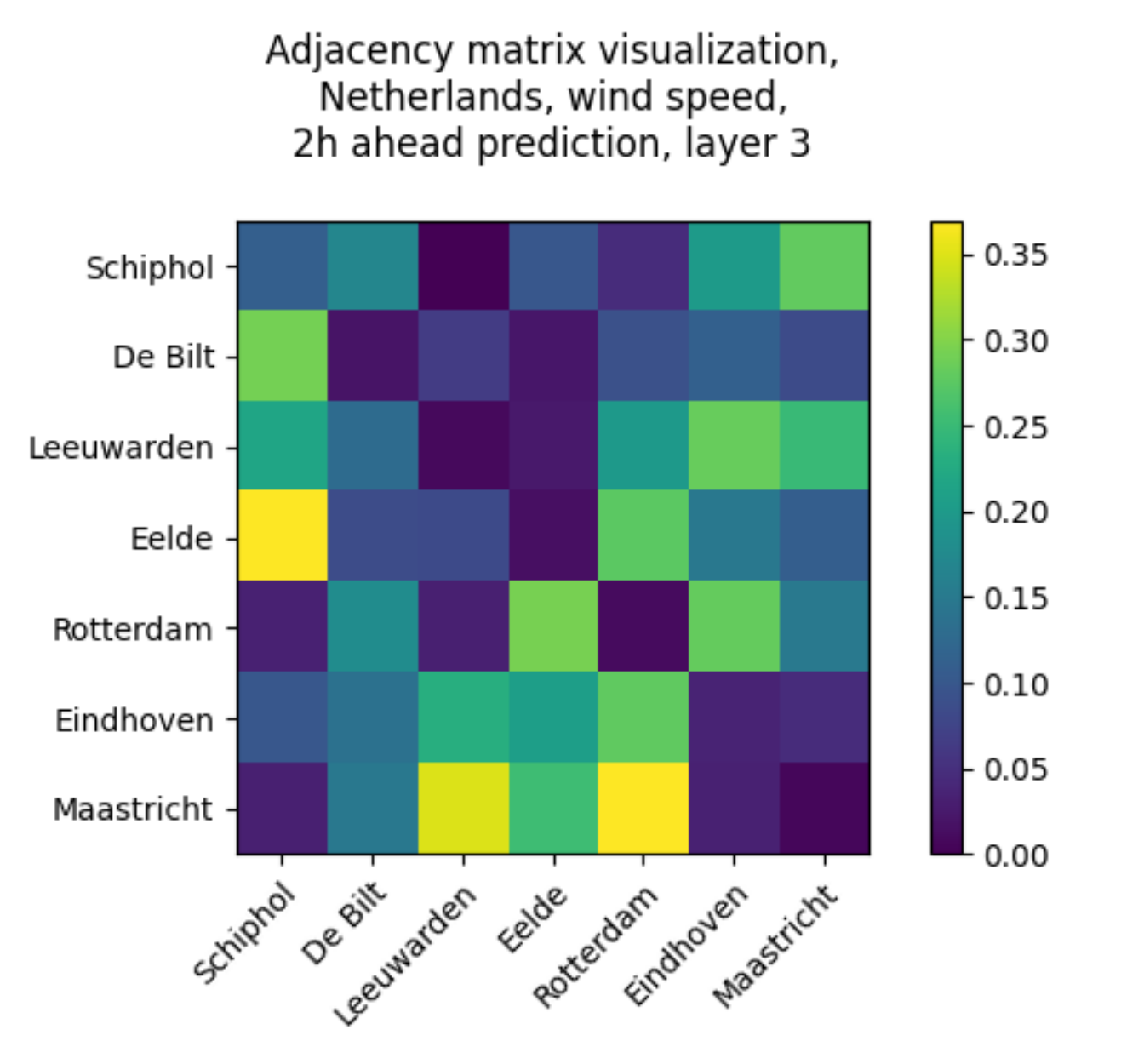}}
    % \\\vskip 0.5pt plus 0.25fil
    \caption{Adjacency matrix visualization for wind speed 2h ahead prediction on Dutch cities dataset. (a,c,e): The visualizations for WeatherGCNet. (b,d,f): The visualizations for WeatherGCNet with $\gamma$. Here, layers 1, 2 and 3 of the models are presented respectively.}
    \label{fig:adj_mat_wind_2h}
\end{figure*}

We also visualize the learnt (and transformed as described in Section \ref{section_methods}) adjacency matrices of the proposed models for the 2h ahead wind speed prediction for the Dutch cities dataset. In particular, Fig. \ref{fig:adj_mat_wind_2h}(a,c,e) show the visualization of the adjacency matrix in first, second and third spatio-temporal layer of WeatherGCNet respectively. Fig. \ref{fig:adj_mat_wind_2h}(b,d,f) show analogous visualization, but for WeatherGCNet with $\gamma$. Fig. \ref{fig:adj_mat_wind_2h}(a,c,e) correspond to the adjacency matrix when we are enforcing the self-connection. Fig. \ref{fig:adj_mat_wind_2h}(b,d,f) correspond to the adjacency matrix when we allow the network to actually decide about the self-connection strength. Given the possibility to decide via the additional parameter $\gamma$, the network prefers the self-loop connections to be relatively small (compared to the other entries), as visible in Fig. \ref{fig:adj_mat_wind_2h}(b,d,f).

The multiplication with adjacency matrix (Section \ref{subsection_spatial_conv}, Eq. \ref{eqn:adjmatmul}) is the only mechanism in the whole model incorporating the information from multiple (in fact from all) graph nodes. All the other operations, e.g. temporal convolution, residual connection process the data independently for each node.
Each entry of adjacency matrix controls how much of information coming from one node is to be included in the information constituting another node, considering all node attributes. Therefore, it can be perceived that the adjacency matrix determines relation between the nodes of the graph (the weather stations) based on the stored attributes - combinations of all features at all considered time steps. 
We denote the learnt and transformed adjacency matrix as $A_{t}$. E.g, in Fig.\ref{fig:adj_mat_wind_2h}(a), presenting the adjacency matrix for wind speed prediction for the Dutch weather stations 2h ahead (WeatherGCNet, layer 1), $A_{t}(7,5)$ shows graphically how much (relatively to other relations) overall information coming from Maastricht is to be included in the resulting information of Rotterdam
in layer 1. In Fig.\ref{fig:adj_mat_wind_2h}(a), it can be observed that relatively much information coming from Eelde is to be included in resulting information of Schiphol $A_{t}(4,1)$ and relatively little information coming from Leeuwarden is to be included in resulting information of Eelde $A_{t}(3,4)$, etc.

\section{Conclusion}\label{section_conclusion}
In this paper, new models based on GCN architecture are proposed for wind speed prediction using
historical weather data. A spatial-temporal graph is designed whose nodes are weather stations and the node attributes are weather variables over time. Thanks to the applied spatial-temporal convolutional operations on the built graph,
the network learns the underlying relations between weather stations through a learnable adjacency matrix of the graph. 
The self-loop connection of the adjacency matrix is enforced with and without learnable scalar parameter enabling the network to decide about the strength of the self connectivity in the graph. The obtained results of the two approaches, with and without learnable scalar parameter, are comparable. In the case that the learnable scalar parameter is involved, the network sets the self-loop connection entries to relatively low values compared with other entries of the learnt adjacency matrix. Comparing with previously introduced baselines, our proposed models provide more accurate predictions in all cases examined.

\section*{Acknowledgment}
Simulations were performed with computing resources granted by RWTH Aachen University.
%%%%%%%%%%%%%%%%%%%%%%%%%%%%%%%%%%%%%%%%
%%%%%%%%%%%%%%%%%%%%%%%%%%%%%%%%%%%%%%%%

\bibliography{References}

\typeout{get arXiv to do 4 passes: Label(s) may have changed. Rerun}
\end{document}